\documentclass[10pt,twocolumn,letterpaper]{article}

\usepackage{iccv}
\usepackage{times}
\usepackage{epsfig}
\usepackage{graphicx}
\usepackage{amsmath}
\usepackage{amssymb}
 \usepackage{color}
 \usepackage{xcolor}

\usepackage[pagebackref=true,breaklinks=true,letterpaper=true,colorlinks,bookmarks=false]{hyperref}

 \iccvfinalcopy 

\def\iccvPaperID{1237} 

\definecolor{gray}{rgb}{0.1,0.1,0.5}

\ificcvfinal\fi
\begin{document}


\title{Recurrent Network Models for Human Dynamics}
\author{Katerina Fragkiadaki \quad \quad Sergey Levine \quad \quad Panna Felsen \quad \quad Jitendra Malik\\
University of California, Berkeley\\
Berkeley, CA \\
{\tt\small \{katef,svlevine@eecs,panna@eecs,malik@eecs\}.berkeley.edu}
}

\pdfoutput=1

\maketitle
\newcommand{\mocap}{\theta}
\newcommand{\ea}{\text{\textit{et al.}}}
\newcommand{\z}{\mathrm{z}}
\newcommand{\tr}{\mathrm{tr}}
\newcommand{\mycolor}{gray}
\newcommand{\ntr}{n_T}
\newcommand{\auto}{\text{ERD}}
\newcommand{\ERD}{\text{ERD}}
\newcommand{\data}{x}
\newcommand{\seq}{\mathbf{x}}
\newcommand{\WW}{\mathbf{W}}
\newcommand{\xex}{\mathbf{x}}
\newcommand{\Lo}{\mathcal{L}}
\newcommand{\thetanet}{\boldsymbol \theta}
\newcommand{\LL}{\mathbf{L}}
\newcommand{\clcl}{\mathrm{cl}^2}
\newcommand{\dl}{\mathrm{dl}}
\newcommand{\ov}{s}
\newcommand{\ATsteer}{\W^{\mathrm{steer}}_{T}}

\newcommand{\Csteer}{\C^{\mathrm{steer}}}
\newcommand{\ndl}{n_D}
\newcommand{\np}{n_P}
\newcommand{\salgroup}{\mathrm{ncut}}
\newcommand{\cnfd}{\mathrm{confidence}}
\newcommand{\thresh}{\mathrm{th}}
\newcommand{\dr}{\mathrm{r}}
\newcommand{\dc}{\ce}
\newcommand{\confcf}{\mathbf{confidence}}
\newcommand{\aligncf}{\mathrm{align}}
\newcommand{\AD}{\mathbf{A}_D}
\newcommand{\W}{\mathbf{W}}
\newcommand{\eig}{\mathrm{eig}}
\newcommand{\pbth}{\beta}
\newcommand{\discthresh}{\gamma}
\newcommand{\Asteer}{\W^\mathrm{steer}_T}
\newcommand{\Asteernoh}{\mathbf{A}_\mathbf{steer}}
\newcommand{\Aco}{\mathbf{A}^{\mathrm{c}}}
\newcommand{\RD}{\mathbf{R}_D}
\newcommand{\dlsel}{h}
\newcommand{\RDF}{\mathbf{R}_D^\dlsel}
\newcommand{\R}{\mathcal{R}}
\newcommand{\AT}{\mathbf{A}}
\newcommand{\f}{\mathrm{f}}
\newcommand{\trc}{\mathrm{C}}
\newcommand{\RT}{\mathbf{R}_T}
\newcommand{\Tf}{\mathcal{T}^\mathbf{f}}
\newcommand{\st}{\mathrm{subject \text{ } to}}
\newcommand{\vc}{\mathrm{vec}}
\newcommand{\Wtilde}{\tilde{\mathbf{W}}}
\newcommand{\thr}{\mathrm{th}}
\newcommand{\maxmarg}{\mathrm{mxmr}}
\newcommand{\trd}{\tr^D}
\newcommand{\C}{\mathbf{C}}
\newcommand{\ns}{n_S}
\newcommand{\nss}{m}
\newcommand{\Cg}{\mathbf{C}_G}
\newcommand{\Prob}{\mathrm{P}}
\newcommand{\D}{\mathbf{D}}

\newcommand{\g}{\mathrm{G}}

\newcommand{\score}{c}

\newcommand{\bbox}{\mathrm{box}}
\newcommand{\X}{\mathrm{X}}
\newcommand{\T}{\mathcal{T}}
\newcommand{\diag}{\mathrm{Diag}}
\newcommand{\Dcap}{\mathcal{D}}
\definecolor{lightgray}{gray}{0.9}
\newcommand{\todo}{\textcolor{red}{TODO: }}
\newcommand{\velocity}{\mathrm{vel}}
\newcommand{\disparity}{\mathrm{dsp}}
\newcommand{\trajdist}{\mathrm{dst}}
\newcommand{\Gup}{\mathrm{G}}
\newcommand{\detection}{R}

\newcommand{\na}{n(\allX)}
\newcommand{\location}{\mathrm {loc}}

\newcommand{\Tr}{\mathrm {Tr}}
\newcommand{\YY}{\mathcal{Y}}
\newcommand{\sel}{\mathbf {s}}
\newcommand{\yf}{y}
\newcommand{\ATDh}{\mathbf{A}_T^{\yf}}
\newcommand{\Ch}{\mathrm{steer}(\C;h)}
\newcommand{\RTh}{\mathbf{R}_T^{\yf}}
\newcommand{\AR}{\mathbf{A}_R}
\newcommand{\pn}{\mathrm{Partition}}
\newcommand{\ptwisemul}{\bullet}

\newcommand{\modify}{M}
\newcommand{\zeroone}{\{0, 1\}}
\newcommand{\ones}{\mathbf 1}
\newcommand{\suchthat}{\mathrm{s.t.}}
\newcommand{\A}{\mathbf{A}}
\newcommand{\GF}{\mathbb{G}}
\newcommand{\e}{e}
\newcommand{\nmultseg}{|\mathrm{Seg}|}
\newcommand{\cut}{\mathrm{cut}}
\newcommand{\degree}{\mathrm{degree}}
\newcommand{\npool}{n(\allX)}
\newcommand{\gscal}{\pbth}
\newcommand{\Dr}{\mathbf{D}}
\newcommand{\allX}{\mathbf X}
\newcommand{\allY}{\mathbf Y}
\newcommand{\footnoteremember}[2]{
\footnote{#2}
\newcounter{#1}
\setcounter{#1}{\value{footnote}}
}
\newcommand{\footnoterecall}[1]{
}


\newcommand{\p}{\mathbf{p}}
\newcommand{\link}{\mathrm{links}}
\newcommand{\vol}{\mathrm{vol}}
\newcommand{\ncut}{\mathrm{ncut}}
\newcommand{\AF}{\mathbb{A}}
\newcommand{\BF}{\mathbb{B}}
\newcommand{\VF}{\mathbb{V}}
\newcommand{\px}{\mathrm{p}}
\newcommand{\Edge}{E^D}
\newcommand{\nr}{n_R}
\newcommand{\xx}{\mathbf{x}}
\newcommand{\ww}{\mathbf{w}}
\newcommand{\wwp}{\mathbf{w}^{D}}

\newcommand{\aex}{a}
\newcommand{\ca}{c}
 \newcommand{\nd}{n_D}

 \newcommand{\Wbar}{\mathbf{\hat{W}}}
 \newcommand{\RSPX}{\mathbf{R}_R}
%

%
%
\newcommand{\DD}{\mathcal D}

%
\newcommand{\nS}{n_{S}}
\newcommand{\den}{\mathrm{\rho}}
\newcommand{\spx}{\mathrm{r}}
\newcommand{\prt}{\mathrm{d}}
\newcommand{\Seg}{\mathbf{S}}
\newcommand{\aff}{\ww}
\newcommand{\affr}{\ww^R}
\newcommand{\Diag}{\mathrm{Diag}}
\newcommand{\conf}{\mathbf{c}}
\newcommand{\stab}{\mathbf{ncut}}
\newcommand{\cmtb}{\mathbf{cmtb}}
\newcommand{\pset}{\mathcal{P}}

\newcommand{\attr}{\mathrm{att}}

%
%
\newcommand{\V}{\mathbf{V}}
\newcommand{\Y}{\mathrm{Y}}

\newcommand{\rr}{\mathrm{r}}
\newcommand{\pa}{p}
\newcommand{\EV}{\mathbf{\Lambda}}
\newcommand{\WE}{\mathbf{W}}
\newcommand{\K}{K}
\newcommand{\ce}{\mathbf{c}}
\newcommand{\width}{\mathrm{w}}
\newcommand{\height}{\mathrm{h}}
\newcommand{\nf}{n_F}
\newcommand{\sss}{\tilde{s}}
\newcommand{\PS}{\mathbf{P}}
\newcommand{\SSp}{\mathcal{S}}
\newcommand{\N}{\mathcal{N}}
\newcommand{\DT}{G}
\newcommand{\di}{\mathbf{d}}
\newcommand{\edge}{\mathrm{e}}
 \newcommand{\ASPX}{\mathbf{A}_{R}}
 \newcommand{\ASPXst}{\mathbf{W}^{\mathrm{steer}}}
 \newcommand{\PP}{\mathcal{D}}
 \newcommand{\bo}{t_f}
%
%

%
 \newcommand{\s}{\mathcal S (\mathcal D)}
%

  \begin{abstract}
We propose the Encoder-Recurrent-Decoder (ERD) model for recognition and prediction of human body pose in videos and motion capture. The ERD model is a  recurrent neural network that incorporates nonlinear encoder and decoder networks before and after  recurrent layers. 
We test instantiations of ERD architectures in the tasks of motion capture (mocap) generation, body pose labeling and body pose forecasting in videos. Our model handles mocap training data across multiple subjects and activity domains, and  synthesizes novel motions while avoiding drifting for long periods of time.   
 For human pose labeling, ERD outperforms a per frame body part detector by resolving left-right body part confusions. 
For video pose forecasting, ERD  predicts body joint displacements across a temporal horizon of 400ms and outperforms a first order motion model based on optical flow. 
ERDs extend previous  Long Short Term Memory (LSTM) models  in the literature to jointly learn representations and their dynamics. Our experiments show such representation learning is crucial for both labeling and prediction in space-time. We find this is a distinguishing feature between the spatio-temporal visual domain in comparison to  1D text, speech or  handwriting, where straightforward hard coded representations have shown excellent results when directly combined with recurrent units \cite{DBLP:conf/nips/SutskeverVL14} .   

\end{abstract}

%

\section{Introduction}

Humans have a remarkable ability to make accurate short-term predictions about the world around them conditioned on prior events \cite{DBLP:series/csm/VernonHF11}. Predicting the movements of other humans is an important facet of these predictions. 
Although the number of possible movements is enormous, conditioning on visual history can reduce the range of probable outcomes to a manageable degree of variation. For example, a walking pedestrian will most likely continue walking, and will probably not begin dancing spontaneously.
Short term predictions of human 
kinematics  
allows people to adjust their behavior, plan their actions, and properly direct their attention when interacting with others.  Similarly, for Computer Vision algorithms, predicting human motion is important for timely human-computer interaction \cite{DBLP:conf/iros/KoppulaS13}, obstacle avoidance \cite{DBLP:conf/iccv/PellegriniESG09}, and people tracking \cite{planningtracking}. While simpler physical phenomena, such as the motion of inanimate objects, can be predicted using known physical laws, there is no simple equation that governs the conscious movements of a person. Predicting the motion of humans instead calls for a statistical approach that can model the range of variation of future behavior, and presents a tremendous challenge for machine learning algorithms.

We address this challenge by introducing Encoder-Recurrent-Decoder (ERD) networks, a type of Recurrent Neural  Network (RNN) model \cite{Williams:1989:LAC:1351124.1351135, Elman} that combines representation learning with learning temporal dynamics. We apply this model to generation, labeling, and forecasting of human kinematics. 
We consider two data domains: motion capture (``mocap'') and video sequences.
For mocap, conditioning on a mocap sequence so far, we learn a distribution over mocap feature vectors in the subsequent frame. At test time, by supplying mocap samples as input back to the model, long  sequences are synthesized. 
For video, conditioning on  a person  bounding box sequence, we predict the body joint locations in the current frame or, for the task of body pose forecasting, at a specific point in the future.
In the mocap case, the input and output domains coincide (3D body joint angles).  
In the  video case, the input and output domains differ (raw video pixels versus body joint locations). 


RNNs  
are network models that process sequential data using recurrent connections between their neural activations at consecutive time steps.  
They have been successfully applied in the language domain for text and handwriting generation  \cite{DBLP:conf/interspeech/KombrinkMKB11,ICML2011Sutskever_524,DBLP:journals/corr/Graves13}, image captioning \cite{DBLP:journals/corr/VinyalsTBE14}, action recognition \cite{DBLP:journals/corr/DonahueHGRVSD14}.  
Ranzato $\ea$ \cite{DBLP:journals/corr/RanzatoSBMCC14}  applies  RNNs for visual prediction by  quantizing the visual signal into a vocabulary of visual words, and predicts a distribution over those words in the next frame, given the visual word sequence observed at a particular  pixel location. 

We advocate a visual predictive model that is ``Lagrangian" in nature \cite{wiki}:  it predicts future outcomes conditioning on an object tracklet rather than on a tube fixated at a particular pixel location, as \cite{DBLP:journals/corr/RanzatoSBMCC14} (the ``Eulerian" approach).  Such object-centric conditioning exploits more relevant visual history of the object for prediction. In contrast, a visual tube fixated at a particular pixel location   encounters dramatically different content under camera or object motion.



%

In the $\auto$, 
the encoder  transforms the input  data to a representation where    learning of  dynamics is easy. The decoder transcribes the output of the recurrent layers to the desired visual form.  For mocap generation, the encoder and decoder are multilayer fully connected networks. For video pose labeling and prediction, the encoder is a Convolutional Neural Network (CNN) \cite{Cun90handwrittendigit} initialized by a CNN per frame body part detector and decoder is a fully connected network.  $\auto$s   simultaneously learn both the representation most suitable for recognition or prediction (input to the recurrent layer), as well as its dynamics, represented in the recurrent weights, by jointly training encoding, decoding and recurrent subnetworks. 
We found such joint finetuning crucial for empirical performance.


\begin{figure*}[ht!]
\begin{center}
\includegraphics[trim=0in 0.5in 0in 0in, scale=0.45]{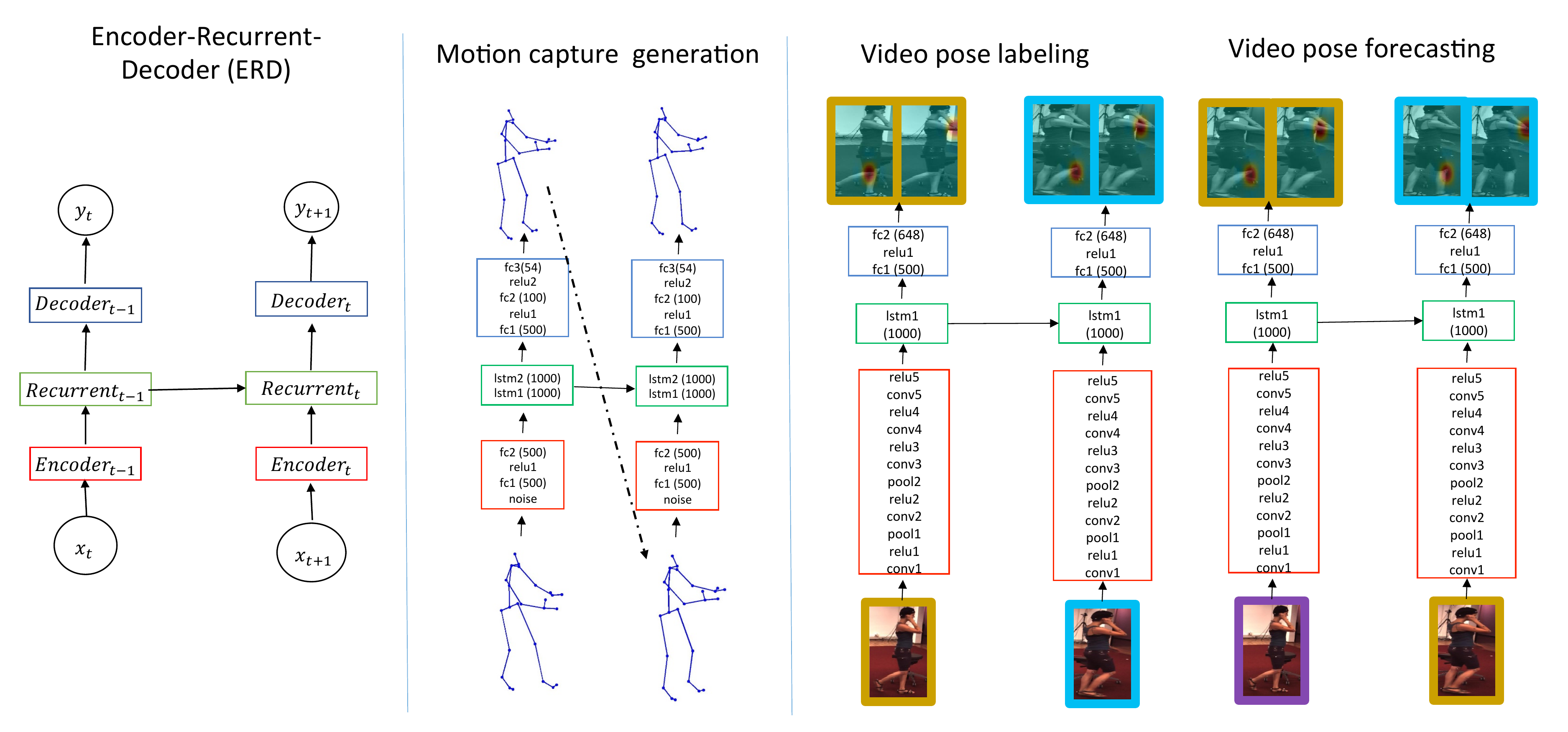}
\end{center}
\caption{
\textbf{ERDs for human dynamics in video and motion capture.}  Given a mocap sequence till time $t$, the $\auto$  for mocap generation predicts the mocap vector at  time instance $t+1$.  
Given a person tracklet till time $t$, $\auto$  for video  forecasting  predicts body joint heat maps  of the \textit{next} frame $t+1$.  $\auto$ for video  labeling predicts heat maps of the \textit{current} frame instead.  
}
\label{fig:autornn}
\end{figure*}

We test $\auto$s in kinematic tracking and forecasting in the H3.6M video pose dataset  of Ionescu $\ea$ \cite{h36m_pami}. It is currently the largest video pose dataset publicly available. It contains a diverse range of activities performed by professional actors and recorded with a Vicon motion capture system. We show that $\auto$s effectively learn human dynamics in video and motion capture. In motion generation, $\auto$s 
synthesize mocap  data across multiple activities and subjects. 
We demonstrate the importance of the nonlinear encoder and decoder in $\auto$s by comparing to previous multilayer LSTM  models \cite{DBLP:journals/corr/Graves13}. We show that such models do not produce realistic motion beyond very short horizons. 
For video pose labeling,  $\auto$s outperforms a per frame body part CNN detector, particularly in the case of left-right confusions.  For future pose forecasting,  $\auto$s   forecast joint positions 400ms in the future, outperforming first order motion modeling with optical flow constancy assumptions. 

Our experiments show that the proposed $\auto$  models can simultaneously model multiple activity domains,  implicitly detect the right activity scenario, and adapt their output labels or predictions accordingly. The action transitioning is  transparent,  in contrast to previous switching dynamical systems or switching HMMs \cite{prm-lslmh-00,Fox:IEEESPM2010} for activity modeling.

\section{Related work} \label{sec:related}

\paragraph{Motion generation}

Generation of naturalistic human motion using probabilistic models trained on motion capture data has previous been addressed in the context of computer graphics and machine learning. Prior work has tackled synthesis of stylized human motion using bilinear spatiotemporal basis models \cite{Akhter:2012:BilinearBasis},  Hidden Markov Models \cite{bh-sm-00}, linear dynamical systems \cite{prm-lslmh-00},  and Gaussian process latent variable models \cite{wfh-gpdmh-08,ufgpd-tclvm-08},  as well as multilinear variants thereof \cite{hpp-sthm-05,wfh-mgpms-07}. Unlike methods based on Gaussian processes, we use a parametric representation and a simple, scalable supervised training method that makes it practical to train on large datasets.

Dynamical models based on Restricted Boltzmann Machines (RBMs) have been proposed for synthesis and infilling of motion data \cite{thr-mhmub-06,sht-rtrbm-08,th-fcrbm-09,DBLP:conf/cvpr/TaylorSFH10}. While such approaches have the advantage of learning probabilistic models, this also results in a substantially more complex training algorithm and, when multilayer models are used, requires sampling for approximate inference. In contrast, our RNN-based models can be trained with a simple stochastic gradient descent method, and can be evaluated very efficiently at test time with simple feedforward operations. 

\paragraph{Video pose labeling and forecasting}
Temporal context has been exploited in kinematic  tracking using  dynamic programming over multiple per frame body pose hypotheses \cite{DBLP:conf/iccv/ParkR11,Batra:2012:DMS:2403138.2403140}, where unary potentials encore detectors' confidence and pairwise potentials encode temporal smoothness. Optical flow has been used  in \cite{conf/cvpr/SappWT11,Sigal:2012:LPE:2205801.2205829} to adjust the temporal smoothness penalty  across consecutive frames. Optical flow can only estimate the motion of body joints that do not move too fast and do not get occluded or dis-occluded. Moreover, the temporal coupling is again pairwise, not long range. $\auto$s keep track of body parts as they become occluded and disoccluded  by aggregating information  in time across multiple frames, rather than the last frame.

Parametric  temporal filters such as Kalman filtering \cite{Weng:2006:VOT:1223195.1223208},   
HMMs or  Gaussian processes for activity specific dynamics \cite{Urtasun:2006:PTG:1153170.1153448,unified,Sminchisescu05discriminativedensity}  generally use simple, linear dynamics models for  prediction.  Such simple dynamics are only valid within very short temporal horizons, making it difficult to incorporate long range temporal information. Switching dynamic systems or HMMs \cite{prm-lslmh-00,Fox:IEEESPM2010} detect activity transitioning explicitly. In contrast, in $\auto$,  action transitioning  is transparent to the engineer, and also more effective. Moreover, HMM capacity increases linearly with increasing numbers of hidden states, but its parameter count increases quadratically. This makes it difficult to scale such models to large and diverse datasets.  
$\auto$s scale better than previous parametric methods in capturing human dynamics. 
RNNs use distributed representations:  each world ``state'' is represented with the ensemble of hidden activations in the recurrent layer, rather than a single one. Thus,  adding a neural unit quadratically increases the number parameters yet  doubles the representation power - assuming binary  units. 

Standard temporal smoothers or filters are  disconnected from the pose detector and operate on its output, such as the space of body joint locations. This representation discards context information that may be present in the original video. In contrast, $\auto$s learn the  representation suitable for temporal reasoning and can take advantage of visual appearance and context.
   \section{$\auto$s for recurrent kinematic tracking and forecasting}

Figure \ref{fig:autornn} illustrates   $\auto$ models for recurrent kinematic tracking and forecasting.    
At  each time step $t$, vector $\data_t$ of a sequence  $\seq=(\data_1, \cdots, \data_T)$ passes through the encoder, the recurrent layers, and the decoder network, producing the output $y_t$. 
In general, we are interested in estimating some function $f(x)$ of the input $x$ at the current time step, or at some time in the future. For example, in the case of motion capture, we are interested in estimating the mocap vector at the next frame. Since both the input and output consists of mocap vectors, $f$ is the identity transformation, and the desired output at step $t$ is $f(x_{t+1})$.
In case of video pose labeling and forecasting, $f(x)$ denotes body joint locations corresponding to the image in the current bounding box $x$. At step $t$, we are interested in estimating either $f(x_t)$ in the case of labeling, or $f(x_{t+H})$ in the case of forecasting, where $H$ is the forecast horizon.

The units in each recurrent layer  implement the Long Short Term Memory functions \cite{Hochreiter:1997:LSM:1246443.1246450}, where writing, resetting, and reading a value from each recurrent hidden unit is explicitly controlled via gating units, as described by Graves \cite{DBLP:journals/corr/Graves13}. Although LSTMs have four times more parameters than regular RNNs, they  facilitate long term storage of task-relevant data. 
In Computer Vision, LSTMs have been used  so far for image captioning  \cite{DBLP:journals/corr/VinyalsTBE14} and action classification in videos \cite{DBLP:journals/corr/DonahueHGRVSD14}.   

$\auto$ architecture extends prior work on LSTMs by augmenting the model with encoder and decoder networks.
Omitting the encoder and decoder networks and instead using linear mappings between the input, recurrent state, and output 
caused underfitting on all three of our tasks. 
This can be explained by the complexity of the mocap and video input in comparison to the words or pen stroke 2D  locations considered in prior work \cite{DBLP:journals/corr/Graves13}. For example, word embeddings were not crucial for RNNs to do well in text generation or machine translation, and the standard one hot encoding vocabulary representation also showed excellent results \cite{DBLP:conf/nips/SutskeverVL14}.

\subsection{Generating Motion Capture}

Our goal is to predict the mocap vector in the next frame, given a mocap sequence so far. Since the output $y_{t}$ has the same format as  the input $x_{t+1}$, if we can predict $x_{t+1}$, we can ``play'' the motion forward in time to generate a novel mocap sequence by feeding the output at the preceding time step as the input to the current one.



Each mocap  vector consists of a set of 3D body joint angles in a kinematic tree representation. 
We represent the orientation of each joint by an exponential map in the coordinate frame of its parent, corresponding to 3 degrees of freedom per joint. The global position of the body in the x-y plane and the global orientation about the vertical z axis are predicted relative to the previous frame, since each clip has an arbitrary global position. This is similar to the approach taken in previous work \cite{thr-mhmub-06}. We standardize our input by mean subtraction and division by the standard deviation along each dimension.

We consider both deterministic and probabilistic predictions. In the deterministic case, the decoder's output $y_t$ is a single mocap vector.  In this case, we train our model by minimizing the Euclidean loss between target and predicted body joint angles. In the probabilistic case, $y_t$  parametrizes  a Gaussian Mixture Model (GMM) over mocap vectors in the next frame.  We then  minimize the GMM negative log-likelihood during training:
\begin{equation}
 \mathcal{L}(\seq) = - \displaystyle\sum_{t=1}^T\mathrm{log}\mathrm{Pr}(x_{t+1}|y_t)
\end{equation}
We use five mixture components and diagonal covariances. The variances are outputs of  exponential layers to ensure  positivity, and the mixture component probabilities are outputs of a softmax layer, similar to \cite{DBLP:journals/corr/Graves13}. During training, we  pad the variances in each iteration by a fixed amount to ensure they do not collapse around the mixture means.
   Weights are initialized randomly. We experimented with initializing the encoder and decoder networks of the mocap $\auto$ from the (first two layers of) encoder and (last two layers of) decoder of a) a ten layer autoencoder trained for dimensionality reduction of mocap  vectors \cite{citeulike:778023}, 
  b) a ``skip'' autoencoder  trained to reconstruct the  mocap vector in few frames in the future given the current one. In both cases, we did not observe improvement over random weight initialization.  We train our $\auto$ model with stochastic gradient descent and backpropagation through time \cite{Williams95gradient-basedlearning} with momentum and gradient clipping at 25, using the publicly available Caffe package \cite{Jia13caffe} and the LSTM layer  implementation from \cite{DBLP:journals/corr/DonahueHGRVSD14}.

We regularize  our mocap $\auto$ with denoising: we provide   mocap vectors corrupted with zero mean Gaussian noise \cite{Vincent:2010:SDA:1756006.1953039} and have the model  predict the correct, uncorrupted  mocap vector in the next frame. We found it valuable to progressively increase the noise standard deviation, learning from non-corrupted  examples first. This corresponds to a type of curriculum learning. 
At test time, we  run the model forward by feeding the predictions as input to the model in the following time step. Without  denoising, this kind of forward unrolling  suffers from accumulation of small prediction mistakes at each frame, and the model  
quickly falls into unnatural regions of the state space. 
Denoising ensures that corrupted mocap data are shown to the network during training so that it learns to  correct small amounts of drift and stay close to the manifold of natural poses.

\subsection{Labeling and forecasting video pose}
In the previous section, we described how the $\auto$ model can be used to synthesize naturalistic human motion by training on  motion capture datasets. 
In this section, we extend this model to identify human poses directly from pixels in a video. We consider a pose labeling task and a pose forecasting task. In the labeling task, given a bounding box sequence depicting a person, we want to estimate body joint locations for the \textit{current} frame, given the sequence so far.
In the forecasting task, we want to estimate body joint locations for a specific future time instance instead.

 
%
%

We represent $K$ body joint locations as a set of $K$ $N \times N$ heat maps over the person's bounding box, that represent likelihood for each joint to appear in each of the $N^2$ grid locations, similar to \cite{vpsKpsTulsianiM14}.  
Predicting heat maps naturally incorporates uncertainty over  body joint locations, as opposed to predicting body joint pixel coordinates. 
 

Figure \ref{fig:autornn}\textit{right} illustrates our $\auto$ architecture for video pose labeling and forecasting.  The encoder is a five layer convolutional network with architecture similar to Krizhevsky $\ea$ \cite{NIPS2012_0534}. 
Our decoder is a two layer network with fully connected layers interleaved with  rectified linear unit layers.  
 The output of the decoder  is    body joint heat maps over the person bounding box in the current frame for the labeling task, or body joint heat maps at a specified future time instance for the forecasting  task. 

We train both our pose labeler and forecaster $\auto$s under a Euclidean loss between estimated and target heat maps. 
We initialize the weights of the encoder from a   six layer convolutional network trained for per frame body part detection, in which the final CONV6 layer corresponds to the body joint heat maps.  

Empirically, we found it valuable to input to the recurrent layer not the per frame estimated heat maps (CONV6), but rather the preceding CONV5  feature maps.   These feature maps capture rich appearance information, rather than merely  body joint likelihood.  
Rich appearance information  assists the network in discriminating between different actions and pose dynamics without  explicit switching across activity domains, as previous switching dynamical linear systems  or HMMs \cite{prm-lslmh-00}.

We use two  networks on different image scales for our per frame pose detector and $\auto$: one where the output layer  resolution is 6$\times$6 and one that works on double  image size and has output resolution of 12$\times$12. The  heat maps of the coarser scale are upsampled and added to the finer scale to provide the final combined 12$\times$12 heat maps. Multiple scales have shown to be beneficial  for static pose estimation  in \cite{vpsKpsTulsianiM14,DBLP:conf/nips/TompsonJLB14,DBLP:journals/corr/ToshevS13}.



\section{Experiments}
We test our method on the H3.6M dataset of Ionescu $\ea$ \cite{h36m_pami}, which is currently the largest video pose dataset. It consists of 15  activity scenarios, 
performed by seven  different professional  actors and  recorded from four static cameras. For each activity scenario, subject, and camera viewpoint, there are two video sequences, each  between 3000 and 5000 frames. Each activity scenario features rich gestures, pose variations and interesting subactions  performed by the actors. For example, the walking activity includes holding hands, carrying a heavy load, putting hands in the pockets,  looking around etc. 
The activities are recorded using a Vicon motion capture system that tracks markers on actors' body joints and provides  high quality 3D body joint locations. 
2D body joints locations are obtained by projecting the 3D positions onto the image plane using the  known camera calibration and viewpoint. For all our experiments, we treat subject 5 as the test subject and all others as our training subjects.


\paragraph{Motion capture generation}

\begin{figure}[h!]
\begin{center}
\includegraphics[trim=0.0in 4in 0in 0.0in, scale=0.28]{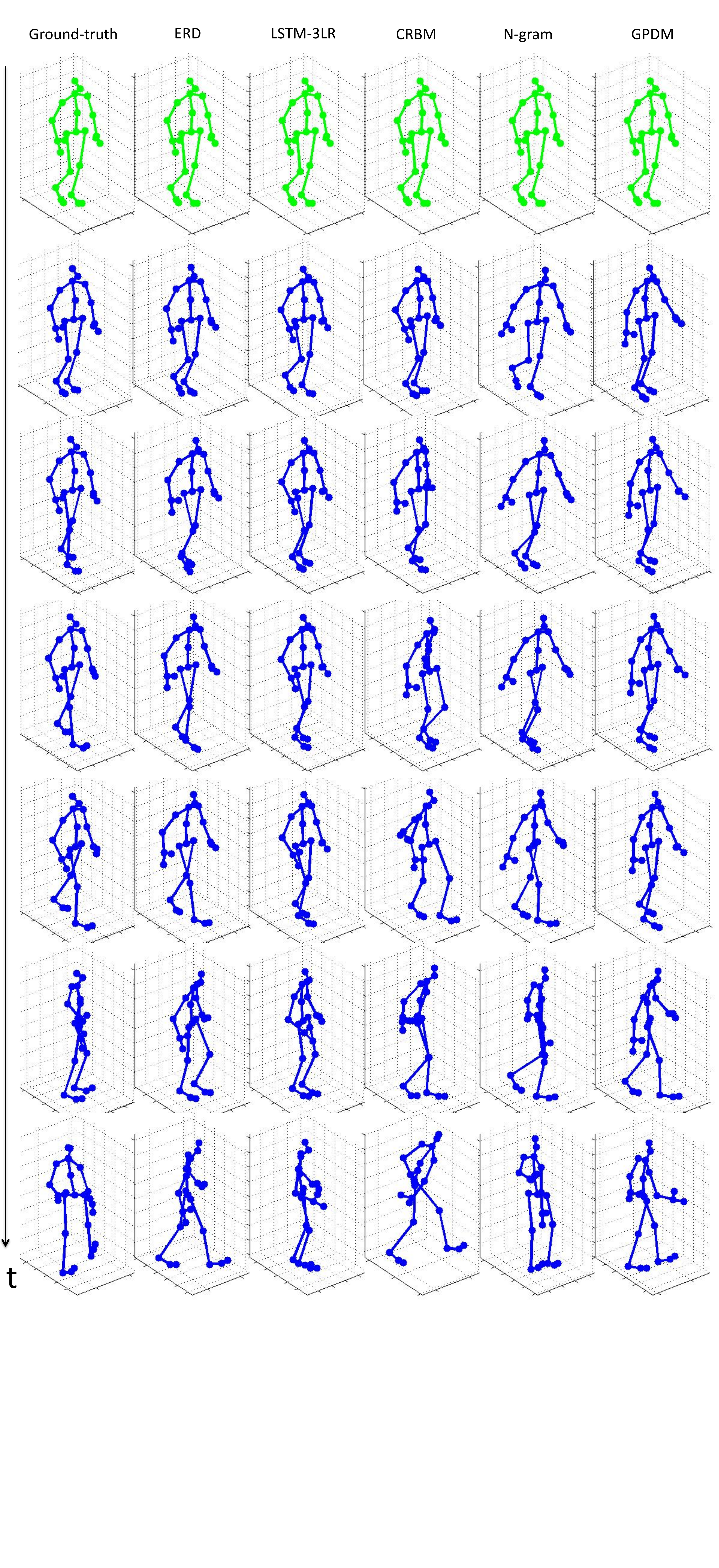}
\end{center}
\caption{ \textbf{Motion synthesis.}  
LSTM-3LR and  CRBMs \cite{thr-mhmub-06} provide  smooth short-term motion completions  (for up to 600msecs), mimicking well novel styles of motion, (e.g., here, walking with upright back).  However,  $\auto$  generates realistic motion for  longer periods of time while LSTM-3LR soon converges to the mean pose and  CRBM  diverges to implausible motion.  NGRAM  has a non-smooth transition from conditioning to generation. Per frame mocap vectors  predicted by GPDM \cite{Wang06gaussianprocess}  look plausible,  but their temporal evolution is far from realistic.  You can watch the corresponding video results  at {\tt https://sites.google.com/site/motionlstm/} }.
\label{fig:crbm}
\end{figure}

We compare our $\auto$ mocap generator with  a) an LSTM recurrent neural network with linear encoder and decoders that has 3 LSTM layers of 1000 units each (architecture found through experimentation to work well), b) Conditional Restricted Boltzmann Machines (CRBMs) of Taylor \textit{et al.} \cite{thr-mhmub-06}, c)  Gaussian Process Dynamic Model (GPDM) of Wang \ea \cite{Wang06gaussianprocess}, and d) a nearest neighbor N-gram model (NGRAM).   For CRBM and GPDM, we used the code made publicly available by the authors. For the nearest neighbor N-gram model, we used a frame window of length $N=6$ and Euclidean distance on 3D angles between the conditioning prefix and our training set,  and copy past the subsequent frames of the best matching training subsequence.  
We applied denoising during training to regularize both the $\auto$ and the LSTM-3LR.  For all models, the mocap frame sequences were subsampled by two. $\auto$, LSTM-3LR and CRBM are trained on multiple activity scenarios (Walking, Eating and Smoking).  GPDM is trained on Walking activity only, because   its  cubic complexity prohibits its training on a large number of  sequences.  
Our comparison focuses  on motion forecasting (prediction) and synthesis, conditioning on motion prefixes of our test subject.  Mocap in-filling and denoising  are nontrivial with our current  model but developing this functionality is an interesting avenue for future work.

We show  qualitative motion synthesis results in Figure \ref{fig:crbm}  and  quantitative motion prediction errors in Table \ref{tab:predictionerror}.  In Figure \ref{fig:crbm},  the conditioning motion prefix from our test subject is shown in green  and the generated motion is shown in blue. 
In Table \ref{tab:predictionerror}, we show Euclidean norm between the synthesized motion  and ground-truth   motion for our test subject  for different temporal horizons past the conditioning motion prefix, the largest being 560msecs, averaged across 8 different prefixes. 
The stochasticity of human motion prevents a metric evaluation for longer temporal horizons, thus all comparisons in previous literature are qualitative.  
LSTM-3LR dominates the short-term motion generation, yet soon converges to the mean pose, as shown in Figure \ref{fig:crbm}. CRBM also provides smooth short term motion completions,  yet quickly drifts to unrealistic motions. $\auto$ provides slightly less smooth completions, yet can generate realistic  motion for long periods of time. For $\auto$, the smallest error was always produced by the most probable GMM sample, which was similar to the output of an ERD trained under a standard Euclidean loss. N-gram model exhibits a sudden change of style during transitioning from the conditioning prefix to the first generated frame, and cannot generate anything outside of the training set. Due to low-dimensional embedding, GPDM cannot adequately handle the breadth of styles in the training data, and produces unrealistic temporal evolution.

The quantitative and qualitative motion generation results of $\auto$ and LSTM-3LR suggest an interesting trade-off between smoothness of motion completion (interesting motion extrapolations) and stable long-term motion generation. Generating short-term motion that mimics the style of the test subject is possible with LSTM-3LR, yet, since the network has not encountered similar  examples during  training, it is unable to correctly generate motion for longer periods of time. In contrast, $\auto$ gears the  generated motion towards similarly moving training examples. $\auto$  though cannot really extrapolate, but rather interpolate among the training subjects. It does   provides much smoother motion completions than the N-gram baseline. Both setups are interesting and useful in different applications, and in between architectures potentially lie somewhere in between the two ends of that spectrum. Finally, it is surprising that LSTM-3LR outperforms CRBMs given its simplicity during  training and testing, not requiring inference over latent variables.

\begin{table} [h]
{\small
\setlength{\tabcolsep}{2.5pt}
\begin{tabular}{|p{1.52cm}||p{0.75cm}|p{0.75cm}|p{0.75cm}|p{0.75cm}|p{0.75cm}|p{0.75cm}|p{0.75cm}|}
\cline{1-8}
                                             & {80}             & {160}       & {240}       &{320}       & {400} & {480} & {560} \\ \hline 
 ERD                                      & 0.89            & 1.39        & 1.93         & 2.38        & 2.76  &  3.09       &  3.41               \\ \hline 
 LSTM-3LR                           & \textbf{0.41} & \textbf{0.67} & \textbf{1.15} & \textbf{1.50} & \textbf{1.78} & \textbf{2.02} & \textbf{2.26}\\   \hline 
CRBM \cite{thr-mhmub-06} & 0.68              &         1.13      & 1.55             & 2.00             & 2.45              & 2.90              & 3.34\\\hline 
6GRAM &                              1.67                & 2.36              & 2.94              & 3.43             & 3.83             &4.19               &4.53  \\\hline 
GPDM \cite{Wang06gaussianprocess} & 1.76 & 2.5 & 3.04 & 3.52 & 3.92 & 4.28 & 4.61\\   \hline 
\end{tabular}
}
 \caption{ \textbf{Motion prediction error} during 80, 160, 240, 320, 400, 480 and 560 msecs past the conditioning prefix for our test subject during Walking activity.  
 Quantitative evaluation for longer temporal horizons is not possible due to stochasticity of human motion. }
\label{tab:predictionerror}
\end{table}

\begin{figure}[ht]
\begin{center}
\includegraphics[trim=0in 0.3in 0in 1in, scale=0.2]{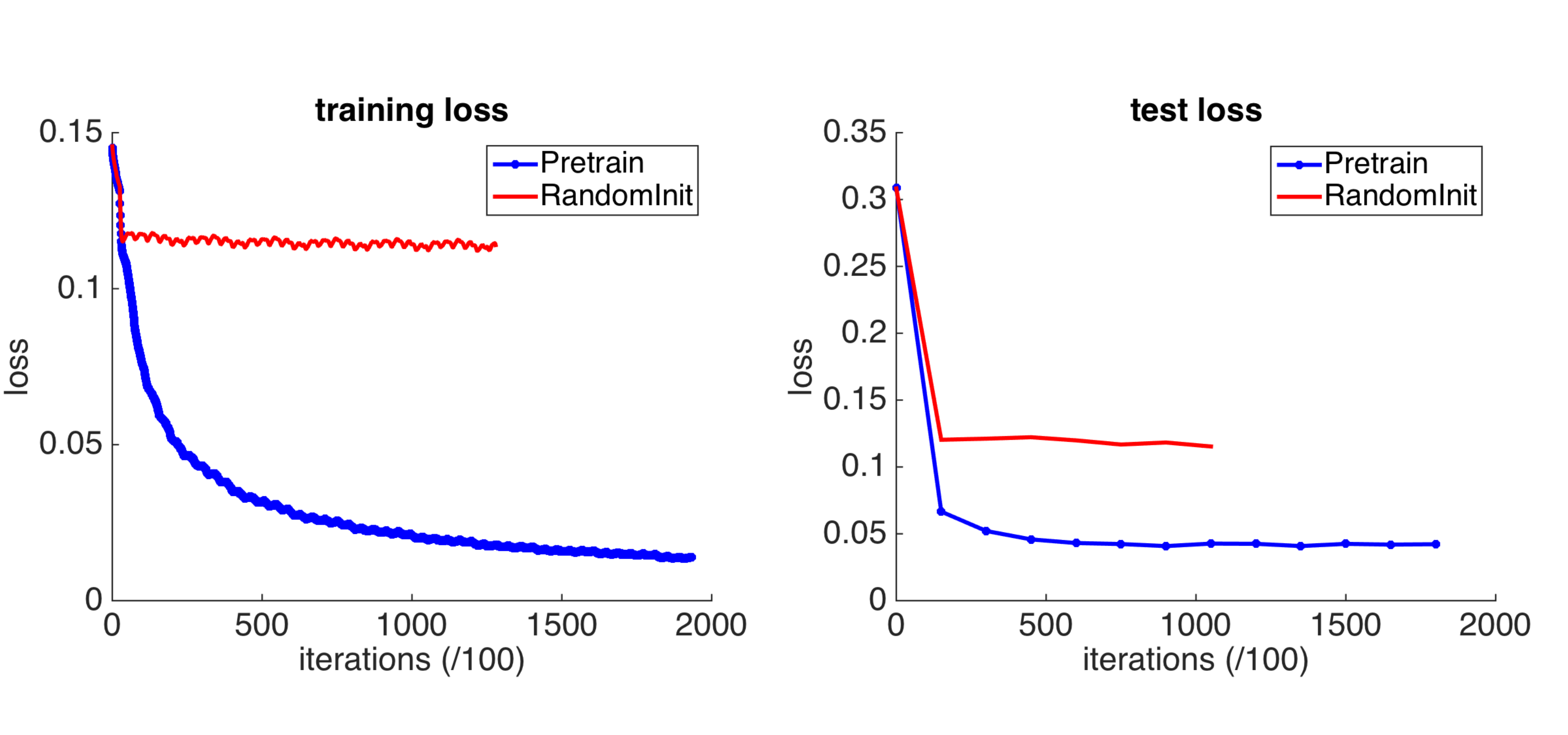}
\end{center}
\caption{\textbf{Pretraining.}  
Initialization of the CNN encoder with the weights of a body pose detector leads to a much better solution than random weight initialization. For motion generation, we did not observe this  performance gap between pertaining and random initialization, potentially due to much shallower encoder and  low dimensionality of the mocap data.}
\label{fig:pretraining}
\end{figure}

\begin{figure*}[ht]
\begin{center}
\includegraphics[trim=0in 0.2in 0in 0in, scale=0.23]{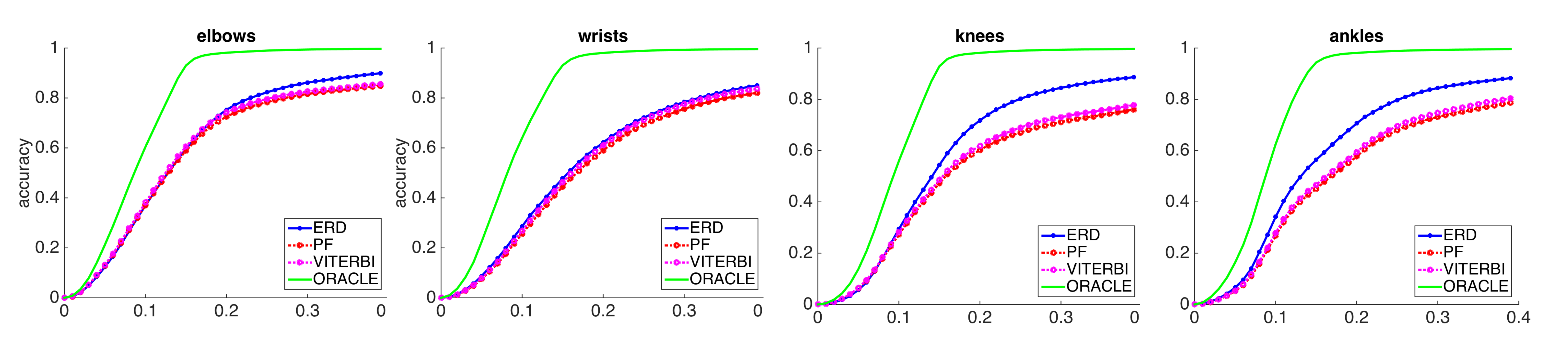}
\end{center}
\caption{\textbf{Video pose labeling in H3.6M.}  Quantitative comparison of  a per frame CNN body part detector  of \cite{vpsKpsTulsianiM14} (PF), dynamic programming for temporal coherence of the body pose sequence in the spirit of \cite{DBLP:conf/iccv/ParkR11,Batra:2012:DMS:2403138.2403140} (VITERBI), and  $\auto$ video pose labeler.  $\auto$  outperforms the per frame detector as well as the dynamic programming baseline.   Oracle curve shows the performance upper-bound  imposed by our grid resolution of 12x12.}
\label{fig:bjtquant}
\end{figure*}

 \begin{figure*}[ht]
 \begin{center}
 \includegraphics[trim=0.3in 0.8in 0in 0.5in, scale=0.67]{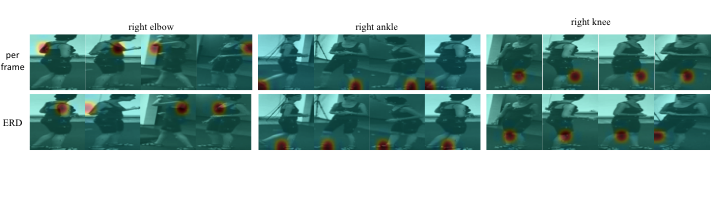}
 \end{center}
 \caption{\textbf{Left-right disambiguation.} $\auto$ corrects  left-right confusions of the per frame CNN detector   by  aggregating   appearance  features (CONV5) across long temporal horizons.}
 \label{fig:leftright}
 \end{figure*}

\paragraph{Video pose labeling}
  Given a person bounding  box sequence, we want to label 2D pixel locations of the person's body joint locations. 
  Both occluded and non-occluded body joints are required to be detected correctly: the occluder's appearance often times contains useful information regarding the location of an occluded body joint \cite{DBLP:conf/eccv/DesaiR12}. 
  Further, for transcribing 2D to 3D pose, all body joints are required  \cite{Taylor:2000:RAO:364058.364079}.

We compare our $\auto$ video labeler against two baselines: a per frame CNN pose detector (PF) used as the encoder part of our $\auto$ model, and a dynamic programming approach over multiple body pose hypotheses per frame (VITERBI) similar in spirit to \cite{DBLP:conf/iccv/ParkR11,Batra:2012:DMS:2403138.2403140}. For our VITERBI baseline, we consider for each body joint in each frame all possible grid locations and encode temporal smoothness as the negative exponential of the Euclidean distance between the locations of the same body joint  across consecutive frames.  The intuition behind VITERBI is that temporal smoothness will help rule out isolated, bad pose estimates, by promoting ones that have lower per frame scores, yet are more temporally coherent.

We evaluate our model and baselines by recording the highest scoring pixel location for each frame and  body joint. We compute the percentage of detected joints within a tolerance radius of a circle centered at the ground-truth body joint locations, for various  tolerance thresholds. We normalize the tolerance radii with the distance between left hip and right shoulder. This is the standard evaluation metric for static image pose labeling \cite{MODEC}. We show pose labeling performance curves  in Figure \ref{fig:bjtquant}. For a video comparison between ERD and the per frame CNN detector, please see the video at { \tt https://sites.google.com/site/motionlstm/}.

discriminatively learning to integrate temporal information  for body joint tracking, instead of employing generic motion smoothness priors.  
 $\auto$'s performance boost stems from correcting left and right confusions of the per frame part detector, as Figure \ref{fig:leftright} qualitatively illustrates.  
Left and right confusion is a major challenge for per frame part detectors, to the extent that certain works measure their performance in image centric coordinates, rather than object centric 
\cite{YangR_CVPR_2011,MODEC}. Last, VITERBI is marginally better than the per frame CNN detector. While motion coherence proved important when combined with shallow and inaccurate per frame body pose detectors \cite{DBLP:conf/iccv/ParkR11,Batra:2012:DMS:2403138.2403140}, it does not improve much upon stronger  multilayer CNNs.

\begin{figure*}[ht]
\begin{center}
\includegraphics[trim=0.0in 1in 0in 0.5in, scale=0.29]{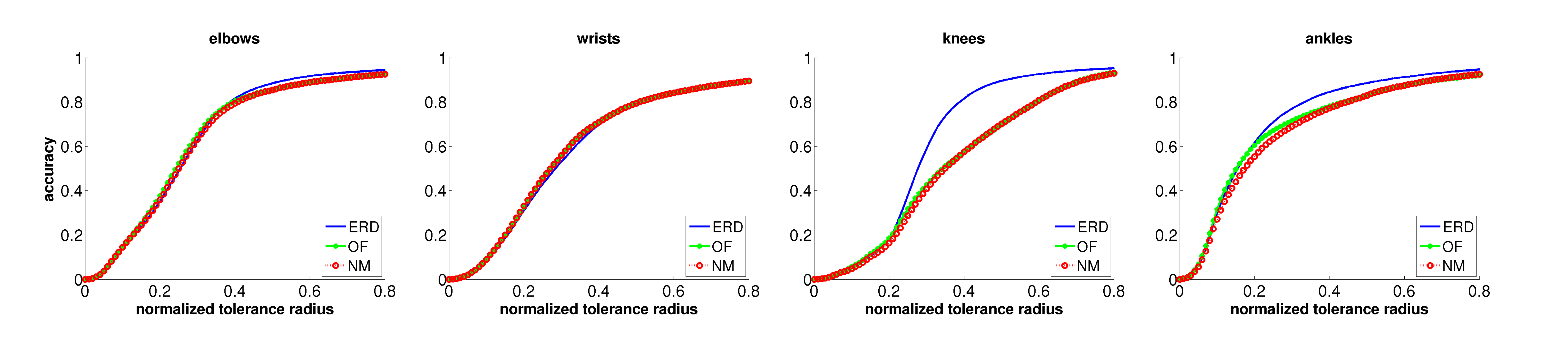}
\end{center}
\caption{ \textbf{Video pose forecasting.} Quantitative comparison between the $\ERD$ model, a zero motion (NM), and constant velocity (OF) models. $\ERD$ outperforms the baselines for the lower body limbs, which are frequently occluded and thus their per frame motion is not frequently observed using optical flow. }
\label{fig:forecasting}
\end{figure*}

Figure \ref{fig:pretraining} compares ERD training and test losses during finetuning  the encoder from (the first five layers of) our  per frame CNN pose detector, versus training the encoder from scratch (random weights). CNN encoder's  initialization is crucial to reach a good solution.  

\begin{figure}[ht]
\begin{center}
\includegraphics[trim=1in 0in 0in 0in, scale=0.2]{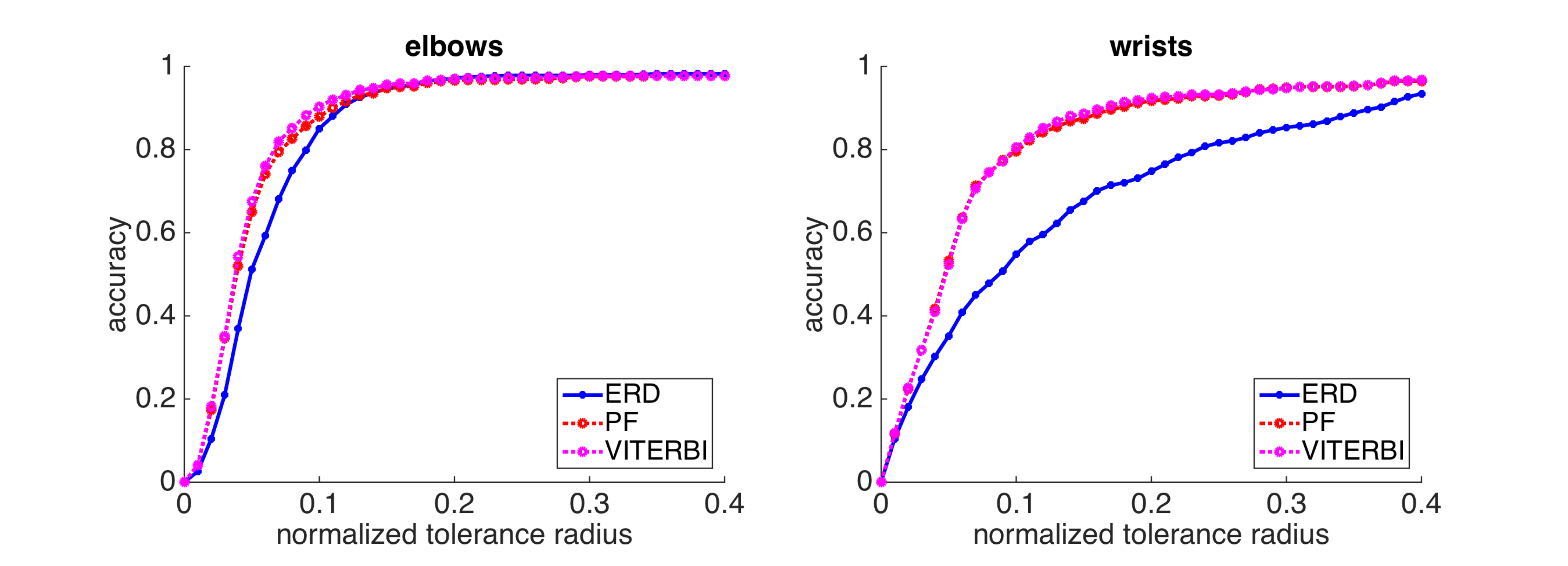}
\end{center}
\caption{\textbf{Video pose labeling in FlicMotion.} 
$\auto$ does not succeed in learning effectively from the small set of 170 videos of about 50 frames each. Large training sets, such as those provided in H3.6M, are necessary for $\auto$ video labeler to outperform generic motion smoothness priors. }
\label{fig:flic}
\end{figure}

We further compare our video labeler in a subset of 200 video sequences of around 50 frames each from the FlicMotion dataset of \cite{DBLP:conf/accv/JainTLB14,MODEC} that we annotated  densely in time with person bounding boxes. We used 170 video sequences for training and 30 for testing. We show performance curves for the upper body joints in Figure \ref{fig:flic}. VITERBI has similar performance as in H3.6M, marginally exceeding the per frame CNN detector. However $\auto$ does much worse since the training set is too small to learn effectively. Finetuning from the model learnt from H3.6M did not help since H3.6M concerns full body motion while FlicMotion captures upper body only.   We did not change the architecture in comparison to the ERD used in H3.6M. It is probable that a smaller recurrent layer and decoder would improve performance preventing overfitting. Large training sets such as in H3.6M allow high capacity discriminative temporal smoothers as our video labelled $\auto$ to outperform generic motion smoothness priors  for human dynamics. 

 \begin{figure}[ht]
 \begin{center}
 \includegraphics[trim=0.0in 0.6in 0in 0in, scale=0.33]{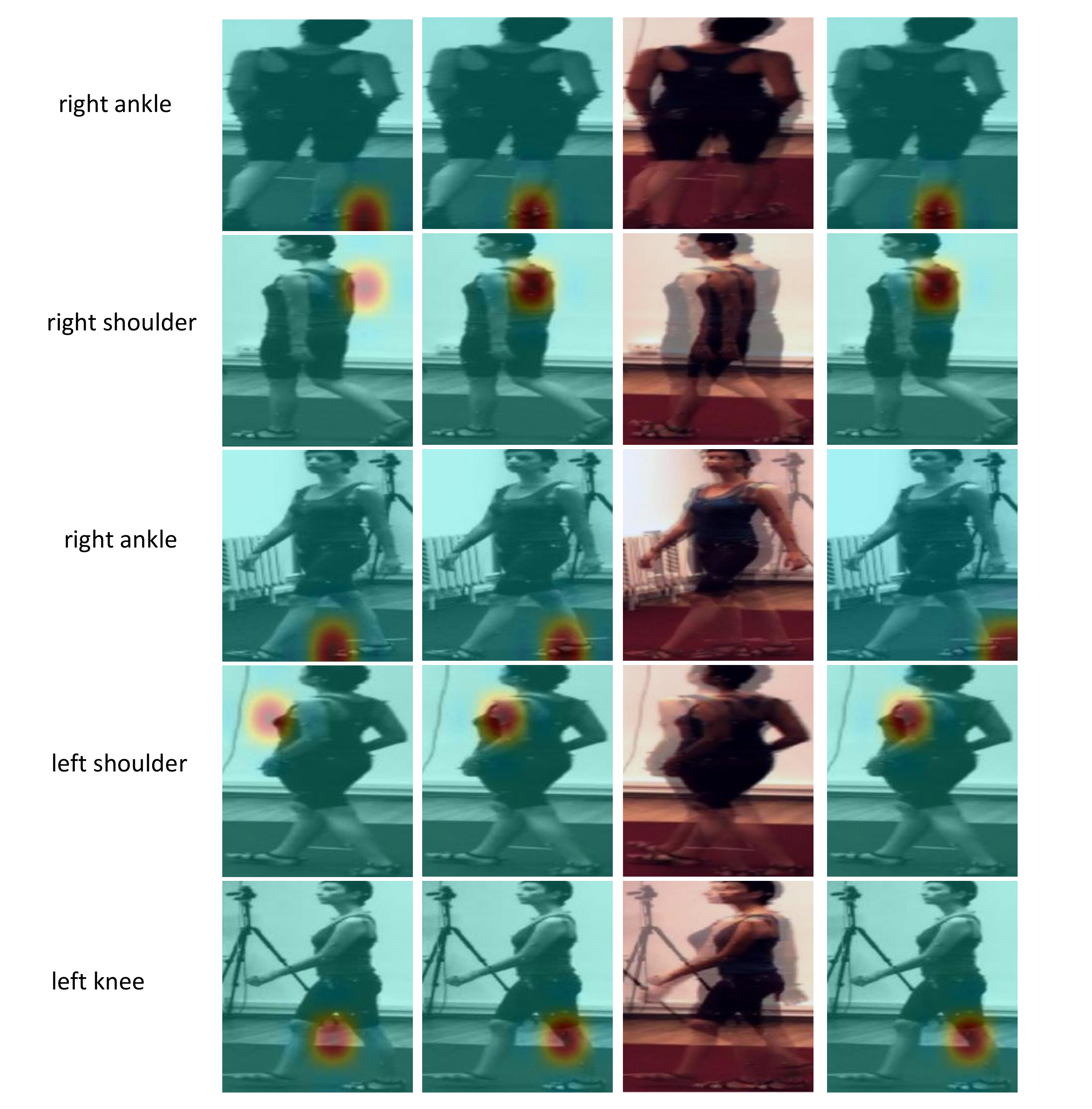}
 \end{center}
 \caption{\textbf{Video pose forecasting}  400ms in the future. \textit{Left:} the prediction of the body part detector 400ms before superimosed on the frame to predict pose for (zero motion model). \textit{MiddleLeft:} Predictions of the $\auto$. The body joints have been moved towards their correct location. \textit{MiddleRight:} The current and 400ms ahead frame superimposed. \textit{Right:} Ground-truth body joint location (discretized in a $N \times N$ heat map grid). In all cases we show the highest scoring heat map grid location. }
 \label{fig:videoprediction}
 \end{figure}

\paragraph{Video pose forecasting}
We predict 2D  body joint locations  400ms ahead of the current frame. 
Figure \ref{fig:forecasting} shows pose forecasting performance curves for our $\ERD$ model, a model that assumes zero object and camera motion  (NoMotion-\textit{NM}), and a model that assumes constant optical flow within the prediction horizon (\textit{OF}).  
$\ERD$ carries out more accurate predictions than the zero order and first order motion baselines, as also shown qualitatively in Figure \ref{fig:videoprediction}.  
Optical flow based motion models cannot make reasonable predictions for occluded body joints, since their frame to frame displacements are not observed.  
Further, standard motion models suffer from separation of the observation model (part detector) and temporal aggregation, which $\auto$ combines into a single network.

\paragraph{Discussion}
Currently, the mocap ERD performs  better on periodic activities (walking, smoking etc) in comparison to non periodic ones (sitting etc.).  Interesting directions for future research is predicting 3D angle differences from frame to frame as opposed to angles directly. Such transformation prediction may generalize better to new subjects, focusing more on  motion rather than  appearance of the skeleton. We are also investigating using large frame volumes as input to our video prediction ERDs with spatio-temporal convolutions in CONV1 as opposed to a single frame LSTM, in order to exploit short temporal horizon more effectively. 

    \section{Conclusion}
We have presented end-to-end  discriminatively trained encoder-recurrent-decoder models for modeling human kinematics in videos and motion capture. $\auto$s   learn the representation for recurrent prediction or labeling, as well as its dynamics, by jointly training encoder recurrent and decoder networks. 
Such expressive models of human dynamics come at a cost of increased need for training examples. In future work, we plan to explore semi-supervised models in this direction, as well learning human dynamics   in multi-person interaction scenarios.

\paragraph{Acknowledgements}
We would like to thank Jeff Donahue and Philipp Kr\"ahenb\"uhl for useful discussions. 
We  gratefully acknowledge NVIDIA corporation for the donation of K40 GPUs for this research. This research was funded by ONR MURI N000014-10-1-0933.

{\small
\bibliographystyle{ieee}
\bibliography{egbib}
}

\end{document}